\documentclass[10pt,conference,a4paper]{IEEEtran}
\usepackage{ifpdf}

\ifCLASSINFOpdf
  \usepackage[pdftex]{graphicx}
  \graphicspath{{../pdf/}{../jpeg/}}
  \DeclareGraphicsExtensions{.pdf,.jpeg,.png}
\else
  \usepackage[dvips]{graphicx}
  \graphicspath{{../eps/}}
  \DeclareGraphicsExtensions{.eps}
  \fi
\usepackage{amsmath,amsxtra,amssymb,amsthm,latexsym,amscd,amsfonts}
\usepackage[utf8]{vntex}
\ifCLASSINFOpdf
\usepackage[pdftex]{graphicx}
\DeclareGraphicsExtensions{.pdf,.jpeg,.png,.jpg}
\else
\usepackage[dvips]{graphicx}

\DeclareGraphicsExtensions{.eps}
\usepackage{multirow}
\usepackage{float}
\usepackage{subcaption}
\fi
\usepackage{array}
\usepackage{epstopdf}
\usepackage{anysize}
\marginsize{2.4cm}{2.0cm}{3.2cm}{3.0cm}
\usepackage{multicol}
\usepackage{balance}
\usepackage{graphicx}
\usepackage{caption}

\makeatletter
\def\ScaleIfNeeded{\ifdim\Gin@nat@width>\linewidth\linewidth\else\Gin@nat@width\fi}

\begin{document}
\columnsep=0.63cm
\def\mathbi#1{\boldsymbol{#1}}
\def\erfc{\:\mathrm{erfc}}
\def\arg{\:\mathrm{arg}}
\def\E{\:\mathrm{E}}
\def\sinc{\:\mathrm{sinc}}
\def\T{\mathrm{T}}
\def\H{\mathrm{H}}
\newcommand{\bigsize}{\fontsize{16pt}{20pt}\selectfont}

%
\include{Abbr}
\title{Theo dõi quỹ đạo Quadrotor sử dụng Linear và Nonlinear Model Predictive Control}

\author{
\IEEEauthorblockN{
Nguyễn Cảnh Thanh, Ngô Huy Hoàng, Đặng Anh Việt và Hoàng Văn Xiêm
} 
\IEEEauthorblockA{ Bộ môn Kỹ thuật Robot, Khoa Điện tử - Viễn Thông \\ Trường Đại học Công Nghệ - Đại học Quốc gia Hà Nội\\
		Email: canhthanhlt@gmail.com, ngoh52180@gmail.com, vietda@vnu.edu.vn, xiemhoang@vnu.edu.vn}
}
\maketitle

\begin{abstract}
Theo dõi quỹ đạo chính xác là đặc tính quan trọng và cần thiết cho điều hướng an toàn của Quadrotor trong trường lộn xộn hoặc bị nhiễu loạn. Trong bài viết này, chúng tôi trình bày chi tiết hai nền tảng điều khiển dựa trên mô hình hiện đại nhất cho bám quỹ đạo: linear-model-predictive controller (LMPC) và nonlinear-model-predictive controller (NMPC). Bên cạnh đó, các mô hình động học, động lực học của quadrotor được mô tả đầy đủ. Cuối cùng, hệ mô phỏng được triển khai và kiểm nghiệm tính khả thi, cho thấy sự hiệu quả của hai bộ điều khiển.

\end{abstract}

\IEEEpeerreviewmaketitle  
\section{GIỚI THIỆU}
\subsection{Bối cảnh và động lực}
Quadrotor đã trở thành một trong những phương tiện bay không người lái (UAV) phổ biến nhất \cite{Wang2016}, góp phần định hình lại các ngành công nghiệp, hậu cần, nông nghiệp,... \cite{Nan2022, Sun2021}. Do cấu trúc đơn giản và khả năng bay linh hoạt, quadrotor dần đóng vai trò quan trọng trọng đời sống cũng như nghiên cứu, tiêu biểu như Hoa Kỳ đã có hơn 300.000 máy bay không người lái thương mại được đăng ký tính đến năm 2021 và theo dự kiến sẽ tiếp tục tăng từ 4,4 tỷ lên 63,6 tỷ USD trong giai đoạn 2018-2025 \cite{Nan2022, Hanover_2022}. Yêu cầu tối quan trọng khi điều khiển quadrotor dưới môi trường nhiễu động bên ngoài có thể ảnh hưởng nặng nề đến hiệu suất bay đặc biệt là ở những khu vực lân cận với các công trình \cite{Mina2016}. Hơn nữa việc áp dụng máy bay không người lái trong trình diễn đặc biệt là trong các sự kiện lớn đòi hỏi cần có khả năng theo đõi đối tượng, quỹ đạo chuyển động một cách nhanh chóng và linh hoạt. Tuy nhiên, hầu hết các phương pháp đều gặp khó khăn trong việc xử lý các hiệu ứng chung trong các chuyến bay nhanh, chẳng hạn như động lực học phi tuyến, hiệu ứng khí động học và giới hạn truyền động \cite{Sun2021}.

Gần đây, model predictive control (MPC) và các biến thể của nó thu hút nhiều sự chú ý cho điều khiển quadrotor nhờ những tiến bộ trong hiệu quả phần cứng và thuật toán \cite{Bicego2020, Foehn_2021}. Ngoài ra, MPC hoạt động đối với các vấn đề đa cảm biến, xem xét các hạn chế đối với đầu vào và trạng thái trong công thức cùng với đặc điểm mạnh mẽ, thích nghi tốt với rối loạn, phi tuyến tính và lỗi mô hình \cite{RIBEIRO201539}.

\subsection{Các nghiên cứu liên quan}

Vấn đề điều khiển của quadrotor đã được nghiên cứu rộng rãi trong đó có nhiều cách tiếp cận khác nhau. Các phương pháp điều khiển tuyến tính như điều khiển proportial-integral-derivative (PID), linear quadratic regulator (LQR) \cite{Khatoon2014, GARCIA2012229} được triển khai với mục đích đạt được tầm bay ổn định và đạt được hiệu suất đủ tốt. Tuy nhiên, phương pháp điều khiển tuyến tính sẽ không còn hiệu quả nếu quỹ đạo là đường đi phức tạp. Các bộ điều khiển phi tuyến được đề xuất tiêu biểu như backstepping \cite{Rashad2015} và feedback linearzation \cite{Zhao2014ANF}.

Phương pháp MPC là một phương pháp điều khiển dựa trên tối ưu hóa, nó tạo ra các lệnh điều khiển theo kiểu đường chân trời rút lui giúp giảm thiểu lỗi theo dòng đường chân trời. Tuy nhiên MPC đòi hỏi rất nhiều về mặt tính toán so với các phương pháp kể trên. Các nghiên cứu linear MPC \cite{BANGURA201411773, Deori2015} được sử dụng trong điều khiển vị trí hoặc điều khiển mô hình tuyến tính hóa. Các nghiên cứu Nonlinear MPC \cite{Mina2016, Nguyen2020} nhằm điều khiển vị trí $xyz$ cũng như vị trí góc ($yaw, pitch, roll$). Các phương pháp Nonlinear MPC cho hiệu quả hoạt động tốt hơn so với Linear MPC.

\subsection{Đóng góp của bài báo}
Đóng góp chính của bài báo là trình bày, so sánh giữa bộ điều khiển linear MPC và nonlinear MPC cho vấn đề theo dõi quỹ đạo đối với robot quadrotor bên cạnh việc mô tả chi tiết mô hình động học, động lực học của robot. Mục đích so sánh là nhấn mạnh lợi ích của việc xem xét mô hình động lực học đối với quỹ đạo chuyển động. Kết quả được mô phỏng và thực hiện trong môi trường MATLAB được hiển thị và xác minh tính hiệu quả của bộ điều khiển MPC.

\subsection{Bố cục bài báo}

Cấu trúc của bài báo được sắp xếp theo thứ tự như sau: Đầu tiên, các vấn đề theo dõi quỹ đạo cùng với mô hình quadrotor được trình bày trong phần  \ref{sec:problemstatement}. Tiếp theo, bộ điều khiển linear MPC và nonlinear MPC được mô tả chi tiết lần lượt trong các phần \ref{Sec:LMPC} và \ref{Sec:NMPC}. Trong phần \ref{Sec:KetQuaMoPhong} đưa ra môi trường giả lập và thảo luận về kết quả bám quỹ đạo của thuật toán. Cuối cùng, kết luận được nêu rõ trong phần \ref{Sec:KetLuan}.
%
%


\section{Phát biểu vấn đề}
\label{sec:problemstatement}

\subsection{Ký hiệu}

\begin{figure}[!ht]
    \centering
    \captionsetup{justification=centering}
    \includegraphics[width=0.5\textwidth]{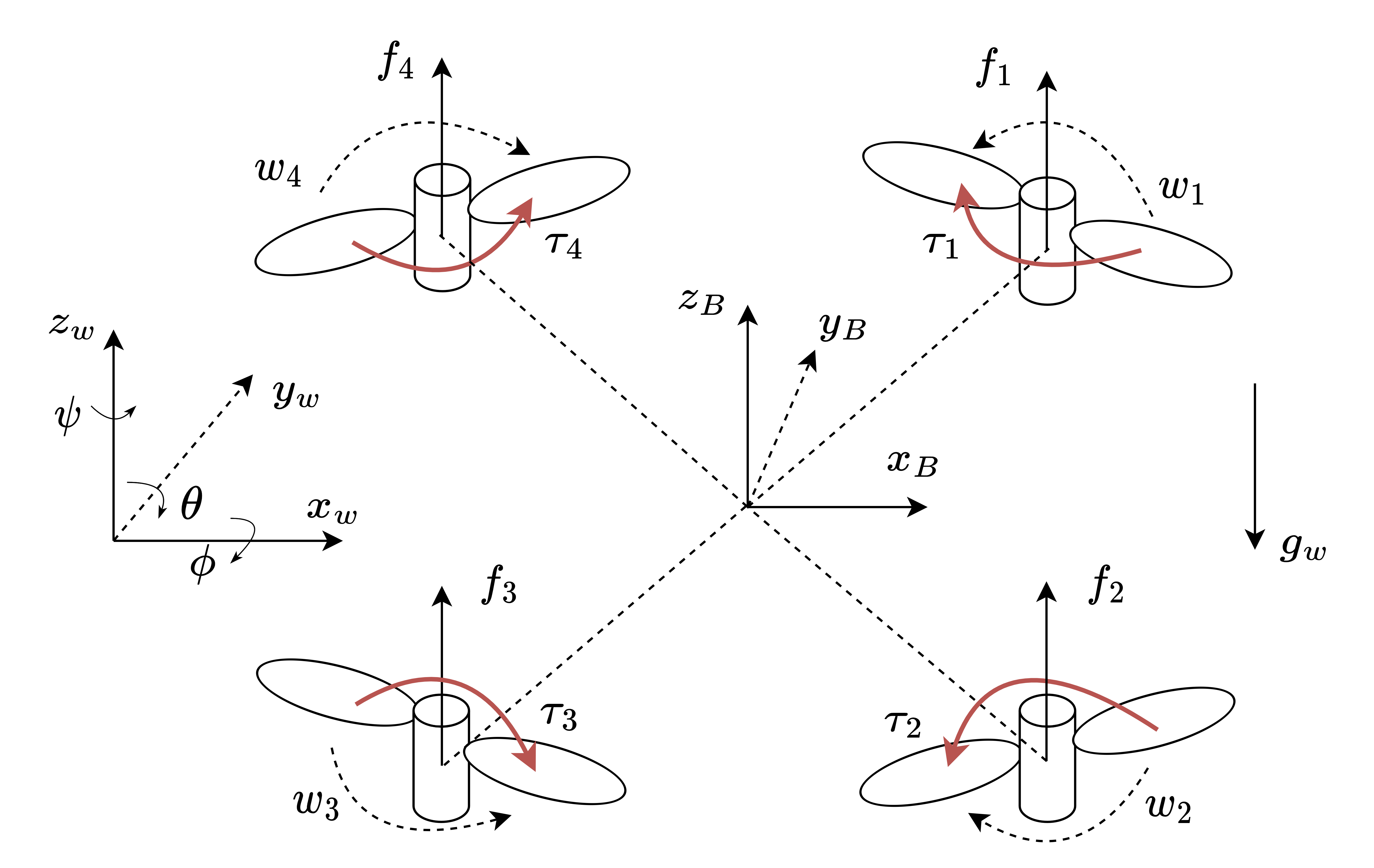}
    \caption{Định nghĩa hệ tọa độ của khung thân và khung quán tính quadrotor}
    \label{fig:quadrotor_model}
\end{figure}
Chúng tôi biểu thị đại lượng vô hướng bằng chữ thường $m$, vector bằng chữ thường in đậm $\mathbi{v}$, ma trận bằng chữ hoa in đậm $\mathbi{R}$. Chúng tôi định nghĩa hai khung tọa độ $W$ - khung tọa độ quán tính, $B$ - khung tọa độ thân như Hình \ref{fig:quadrotor_model} với hệ trực chuẩn như $\{\mathbi{x}_w, \mathbi{y}_w, \mathbi{z}_w\}$. Khung $B$ là vị trí trọng tâm của quadrotor với giả thiết rằng tất cả bốn cánh quạt nằm trong mặt phẳng $xy$ của khung $B$. Vector $\mathbi{\xi} = 
\begin{bmatrix}
    x & y & z
\end{bmatrix}^T$  đại diện cho vị trí tuyệt đối của quadrotor trong khung quán tính $W$. Tiếp theo, vị trí góc được định nghĩa theo vector Euler $\mathbi{\eta} = \begin{bmatrix}
    \phi & \theta & \psi
\end{bmatrix}^T$. Các góc khớp được giới hạn như sau: góc roll $(-\pi/2 < \phi < \pi/2)$, góc pitch $(-\pi/2 < \theta < \pi/2)$, góc yaw $(-\pi < \psi < \pi)$. Chúng tôi viết tắt $S_x = sin(x)$, $C_x = cos(x)$, $T_x = tan(x)$. Vector $\mathbi{q} = \begin{bmatrix}
    \mathbi{\xi} & \mathbi{\eta}
\end{bmatrix} ^ T \in \mathbb{R}^6$ bao gồm cả vị trí góc và vị trí quán tính. Trong khung $B$, vận tốc tuyến tính và vận tốc góc lần lượt được xác định bởi $\mathbi{v}_{B} = \begin{bmatrix}
    v_{xB} & v_{yB} & v_{zB}
\end{bmatrix}^T, \mathbi{w}_{B} = \begin{bmatrix}
    p & q & r
\end{bmatrix}^T.$

\subsection{Mô hình động học}

Ma trận xoay giữa khung $B$ và khung $W$ được định nghĩa theo ba ma trận xoay theo từng hướng:
\begin{equation}
\label{eq:rotation}
    \begin{aligned}
        ^W\mathbi{R}_B &= ^W\mathbi{R}_{zB} \cdot  ^W\mathbi{R}_{yB} \cdot ^W\mathbi{R}_P{xB}  \\ &=
        \begin{bmatrix}
            C_\psi C_\theta & C_\psi S_\theta S_\phi - S_\psi C_\phi & C_\psi S_\theta C_\phi + S_\psi S_\phi \\
            S_\psi C_\theta & S_\psi S_\theta S_\phi + C_\psi C_\phi & S_\psi S_\theta C_\phi - C_\psi S_\phi \\
            -S_\theta & C_\theta S_\phi & C_\theta C_\phi
        \end{bmatrix}.
    \end{aligned}
\end{equation}

Phương trình động học tịnh tiến được định nghĩa như sau:
\begin{equation}
    \begin{aligned}
        \mathbi{v}_W = ^W\mathbi{R}_B \cdot \mathbi{v}_{B},
    \end{aligned}
\end{equation}
trong đó $\mathbi{v}_W = \begin{bmatrix}
    v_x & v_y & v_z
\end{bmatrix}^T $ là vận tốc tuyến tính trong khung $W$.

Phương trình động học xoay được định nghĩa như sau:
\begin{equation}
\begin{array}{*{20}{c}}
    \dot{\mathbi{\eta}} = \mathbi{W}_\eta^{-1} \mathbi{\omega}_B , 
    \begin{bmatrix}
        \dot{\phi} \\ \dot{\theta} \\ \dot{\psi} 
    \end{bmatrix} = 
    \begin{bmatrix}
        1 & S_\phi T_\theta & C_\phi T_\theta \\
        0 & C_\phi & -S_\phi \\
        0 & S_\phi/C_\theta & C_\phi/C_\theta
    \end{bmatrix}
    \begin{bmatrix}
        p \\ q \\ r
    \end{bmatrix}, \\
    \mathbi{\omega}_B = \mathbi{W}_\eta \dot{\mathbi{\eta}} ,
    \begin{bmatrix}
        p \\ q \\ r
    \end{bmatrix}
     = 
    \begin{bmatrix}
        1 & 0 & -S_\theta \\
        0 & C_\phi & C_\theta S_\phi \\
        0 & -S_\phi& C_\theta C_\phi
    \end{bmatrix}
    \begin{bmatrix}
        \dot{\phi} \\ \dot{\theta} \\ \dot{\psi} 
    \end{bmatrix},
    \end{array}
\end{equation}
trong đó $\mathbi{W}_\eta$ được gọi là ma trận Euler biến đổi vận tốc góc từ khung $W$ sang khung $B$ \cite{Saber2001}.

\subsection{Mô hình động lực học}
Vận tốc góc của rotor thứ $i$ ký hiệu là $w_i$ tạo ra lực $f_i$ 
và tổng lực đẩy $\mathbi{T}_B$ theo phương của trục $z$ rotor.
\begin{equation}
    f_i = k \omega^{2}_{i}, T = \sum_{i=1}^{4}f_i = k \sum_{i=1}^{4}\omega^{2}_i, \mathbi{T}_B = \begin{bmatrix}
        0 \\ 0 \\ T
    \end{bmatrix}
\end{equation}

Phương trình Lagrange-Euler biểu diễn phương trình chuyển động dựa trên khái niệm động năng và thế năng bao gồm năng lượng động năng tịnh tiến $E_{trans}$, năng lượng động năng xoay $E_{rot}$ và năng lượng thế năng $E_{pot}$
\begin{equation}
\begin{aligned}
    \mathcal{L}(\mathbi{q, \dot{q})} &= E_{trans} + E_{rot} - E_{pot}, \\
    & = (m/2) \dot{\mathbi{\xi}}^T \dot{\mathbi{\xi}} + (1/2) \mathbi{\omega}^{T}_{B} \mathbi{I} \mathbi{\omega}_B -mgz .
    \end{aligned}
\end{equation}

Như đã được trình bày trong bài \cite{luukkonen2011modelling, Castillo2005}, phương trình Euler-Lagrange với lực và mô-men là:
\begin{equation}
    \begin{bmatrix}
        \mathbi{f}_\xi \\ \mathbi{\tau}_\eta
    \end{bmatrix} = \frac{d}{dt} \bigg(\frac{\partial \mathcal{L}}{\partial \dot{\mathbi{q}}}\bigg) - \frac{\partial \mathcal{L}}{\partial \mathbi{q}} ,
\end{equation}
trong đó $\mathbi{\tau}_\eta \in \mathbb{R}^3$ đại diện cho mô-men roll, pitch, yaw. $\mathbi{f}_\xi$ là lực tịnh tiến tác dụng lên quadrotor. Phương trình Euler-Lagrange tuyến tính như sau:
\begin{equation}
\label{eq:forceeq}
    \mathbi{f}_\xi = ^W\mathbi{R}_B \mathbi{T}_B + \mathbi{\alpha}_T
    = m \ddot{\mathbi{\xi}} + mg \begin{bmatrix}
        0 \\ 0 \\ 1
    \end{bmatrix} + \mathbi{\alpha}_T,
\end{equation}
trong đó $\mathbi{\alpha}_T = \begin{bmatrix}
    A_x & A_y & A_z
\end{bmatrix}^T$ là vector nhiễu động bên ngoài. 
Từ phương trình (\ref{eq:forceeq}) ta có vector trạng thái $\mathbi{\xi}$ như sau:
\begin{equation}
\begin{bmatrix}
    \ddot{x} \\ \ddot{y} \\ \ddot{z}
\end{bmatrix} = -g \begin{bmatrix}
    0 \\ 0 \\ 1
\end{bmatrix} + \frac{T}{m} \begin{bmatrix}
    C_\psi S_\theta C_\phi + S_\psi S_\phi \\
    S_\psi S_\theta C_\phi - C_\psi S_\phi \\
    C_\theta C_\phi
\end{bmatrix} ,
\end{equation}
trong đó $m, g$ lần lượt là trọng lượng của quadrotor và gia tốc trọng trường.

Ma trận Jacobian $\mathbi{J}_\eta$ từ $\mathbi{\omega}_B$ tới $\dot{\mathbi{\eta}}$ biểu thị năng lượng quay trong khung quán tính được định nghĩa như sau:
\begin{align}
    \mathbi{J}_\eta = \mathbi{W}_\eta^T \mathbi{I} \mathbi{W}_\eta
    =  \begin{bmatrix}
        I_{xx} & 0 \\
        0 & I_{yy} C^2_\phi + I_{zz} S^2_\phi \\
        -I_{xx}S_\theta & (I_{yy} - I_{zz})C_\phi S_\phi C_\theta
    \end{bmatrix} \nonumber \\
     \begin{bmatrix}
        & -I_{xx}S_\theta \\
        & (I_{yy} - I_{zz})C_\phi S_\phi C_\theta \\
        & I_{xx} S^2_\theta + I_{yy} S^2_\phi C^2_\theta + I_{zz} C^2_\phi C^2_\theta
    \end{bmatrix},
    \end{align}
trong đó $\mathbi{I} = \begin{bmatrix}
    I_{xx} & 0 & 0 \\
    0 & I_{yy} & 0 \\
    0 & 0 & I_{zz}
\end{bmatrix}$ là ma trận đường chéo trong đó $I_{xx} = I_{yy}$. Sau đó, công thức động học xoay được xác định theo công thức:
\begin{equation}
    E_{rot} = \frac{1}{2}\ddot{\eta}^T \mathbi{J}_\eta \ddot{\eta}.
\end{equation}

Phương trình góc Euler-Lagrange là:
\begin{equation}
    \mathbi{\tau}_\eta = \mathbi{J}_\eta \ddot{\mathbi{\eta}} + \mathbi{C}(\mathbi{\eta}, \dot{\mathbi{\eta}})\dot{\mathbi{\eta}},
\end{equation}
với ma trận $\mathbi{C(\eta, \dot{\eta})}$ là ma trận Coriolis:
\begin{equation}
    \mathbi{C(\eta, \dot{\eta})} = \begin{bmatrix}
        C_{11} & C_{12} & C_{13} \\
        C_{21} & C_{22} & C_{23} \\
        C_{31} & C_{32} & C_{33}
    \end{bmatrix},
\end{equation}
trong đó:

$ C_{11} = 0  $
\begin{multline*}
{ C_{12} =  (I_{yy} - I_{zz}) (\dot{\theta} C_\phi S_\phi + \dot{\psi} S^2_\phi C_\theta)}  \\ 
+(I_{zz} - I_{yy}) \dot{\psi} C^2_\phi C_\theta - I_{xx} \dot{\psi} C_\theta
\end{multline*}

$ C_{13} = (I_{zz} - I_{yy}) \dot{\psi} C_\phi S_\phi C^2_\theta$
\begin{multline*}
C_{21} =  (I_{zz} - I_{yy}) (\dot{\theta} C_\phi S_\phi + \dot{\psi} S_\phi C_\theta)  \\ 
+(I_{yy} - I_{zz}) \dot{\psi} C^2_\phi C_\theta + I_{xx} \dot{\psi} C_\theta
\end{multline*}

$ C_{22} = (I_{zz} - I_{yy}) \dot{\phi} C_\phi S_\phi$

$C_{23} =  -I_{xx} \dot{\psi} S_\theta C_\theta + I_{yy} \dot{\psi} S^2_\phi S_\theta C_\theta +I_{zz} \dot{\psi} C^2_\phi S_\theta C_\theta$

$ C_{31} = (I_{yy} - I_{zz}) \dot{\psi} C^2_\theta S_\phi C_\phi - I_{xx} \dot{\theta} C_\theta$
\begin{multline*}
C_{32} =  (I_{zz} - I_{yy}) (\dot{\theta} C_\phi S_\phi S_\theta + \dot{\phi} S^2_\phi C_\theta) +(I_{yy} - I_{zz}) \dot{\phi} C^2_\phi C_\theta \\ 
+ I_{xx} \dot{\psi} S_\theta C_\theta - I_{yy} \dot{\psi} S^2_\phi S_\theta C_\theta - I_{zz}\dot{\psi} C^2_\phi S_\theta C_\theta 
\end{multline*}
\begin{multline*}
C_{33} =  (I_{yy} - I_{zz}) \dot{\phi} C_\phi S_\phi C^2_\theta - I_{yy} \dot{\theta} S^2_\phi C_\theta S_\theta \\ 
-I_{zz} \dot{\theta} C^2_\phi C_\theta S_\theta + I_{xx} \dot{\theta} C_\theta S_\theta.
\end{multline*}

Do đó, mô hình chuyển động quay của quadrotor được đưa ra như sau:
\begin{equation}
    \ddot{\mathbi{\eta}} = \mathbi{J}_\eta ^{-1} (\mathbi{\tau}_\eta - \mathbi{C}(\mathbi{\eta, \dot{\eta}}) \mathbi{\dot{\eta}} )
\end{equation}

\section{Linear MPC}
\label{Sec:LMPC}
Trong phần này, bài toán linear MPC được xây dựng như một bài toán lập trình bậc hai. Chúng tôi định nghĩa vector trạng thái $\mathbi{x} \in \mathbb{R}^{12}$ như sau:
\begin{equation}
    \mathbi{x} = \{\mathbi{\xi}^T, \mathbi{\eta}, \dot{\mathbi{\xi}}^T, \dot{\mathbi{\eta}}^T\}.
\end{equation}
và vector đầu vào $\mathbi{u} \in \mathbb{R}^{4}$:
\begin{equation}
    \mathbi{u} = \{\omega_1^2, \omega_2^2, \omega_3^2, \omega_4^2\}.
\end{equation}

Sau khi tuyến tính hóa và rời rạc hóa, mô hình không gian trạng thái tuyến tính được xác định theo công thức:
\begin{equation}
    \mathbi{x}_{k+1} = \mathbi{A}\mathbi{x}_k + \mathbi{B}\mathbi{u}_k + \mathbi{V}_d\mathbi{F}_{e,k},
\end{equation}
trong đó $\mathbi{F}_{e,k}$ là các lực bên ngoài, $\mathbi{B}_d$ là ma trận nhiễu loạn.

LMPC có thể được xây dựng như một quy trình tối ưu hóa (OCP) lặp đi lặp lại với giả định rằng các ràng buộc đầu vào được áp dụng nhưng không có ràng buộc trạng thái nào. Hàm chi phí mục tiêu $J(\mathbi{x}, \mathbi{u})$ được định nghĩa như sau:
\begin{equation}
\label{eq:lostfunction}
    J(\mathbi{x}, \mathbi{u}) = \sum_{k = 0}^{N-1}( \|^e\mathbi{x_k} \|^2_{\mathbi{Q}}) + \sum_{k = 0}^{N_u-1}(\|\mathbi{u}_k \|^2_{\mathbi{R}}) +\|^e\mathbi{x}_N\|^2_{\mathbi{P}},
\end{equation}
với:
\begin{equation}
\left\{\begin{matrix}
\mathbi{x}_{k+1} = f(\mathbi{x}_k, \mathbi{u}_k, \mathbi{F}_{e, k})\\ 
\mathbi{F}_{e,k+1} = \mathbi{F}_{e, k} \\
\mathbi{u}_{min} \leqslant \mathbi{u}_k \leqslant \mathbi{u}_{max} \\
\mathbi{x}_0 = \mathbi{x}(t_0), \mathbi{F}_{e, 0} = \mathbi{F}_{e}(t_0)
\end{matrix}\right. \hspace{0.2cm} \forall k \in \left[0, N-1\right],
\end{equation}

trong đó $\| \cdot \|$ biểu thị khoảng cách Euclidean. $\mathbi{Q} \geqslant 
 0$, $\mathbi{R} \geqslant 0$, $ \mathbi{P} \geqslant  0$ lần lượt là ma trận trọng số của trạng thái, đầu vào và trạng thái cuối cùng. $N$ là số bước dự đoán chân trời và $N_u$ là số bước điều khiển chân trời. $\mathbi{u}_{min}$, $\mathbi{u}_{max}$ là giới hạn dưới và giới hạn trên của tín hiệu điều khiển. 

Sai số của theo dõi quỹ đạo được định nghĩa bởi trạng thái hiện tại $\mathbi{x}_k$ và trạng thái tham chiếu $^r \mathbi{x}_k$:
\begin{equation}
     ^e\mathbi{x}_k = \mathbi{x}_k - ^r\mathbi{x}_k \hspace{0.5cm} \forall k 
     \in  \left[0, N\right].
\end{equation}

Trong trường hợp của chúng tôi, sáu trạng thái đầu vào $[ x, y, z, \phi, \theta, \psi]$ phải tuân theo quỹ đạo tham chiếu do số lượng tín hiệu điều khiển nhỏ hơn số lượng tham chiếu đầu ra dẫn tới không đủ bậc tự do để theo dõi quỹ đạo độc lập cho tất cả đầu ra.

Phương trình (\ref{eq:lostfunction}) được viết lại như sau:
\begin{equation}
    J(\bar{\mathbi{x}}, \bar{\mathbi{u}}) = \bar{\mathbi{x}}^T \bar{\mathbi{Q}} \bar{\mathbi{x}} + \bar{\mathbi{u}}^T \bar{\mathbi{R}} \bar{\mathbi{u}} + ^e\mathbi{x}^T_N\mathbi{P} \ ^e\mathbi{x}_N
\end{equation}
trong đó, 

$\bar{\mathbi{x}} = \begin{bmatrix}
    ^e\mathbi{x}_1 & ^e\mathbi{x}_2 & \dots & ^e\mathbi{x}_N
\end{bmatrix} \in \mathbb{R}^{N}$ \\

$\bar{\mathbi{u}} = \begin{bmatrix}
    \mathbi{u}_1 & \mathbi{u}_2 & \dots & \mathbi{u}_{N_u-1}
\end{bmatrix} \in \mathbb{R}^{N_u}$ \\

$\bar{\mathbi{Q}} = \begin{bmatrix}
    Q_1 & 0 & \dots & 0 \\
    0 & Q_2 & \dots & 0 \\
    \vdots  & \vdots & \ddots & 0 \\
    0 & 0 & \dots & Q_N \\
\end{bmatrix}$ \\ \\

$\bar{\mathbi{R}} = \begin{bmatrix}
    R & 0 & \dots & 0 \\
    0 & R & \dots & 0 \\
    \vdots  & \vdots & \ddots & 0 \\
    0 & 0 & \dots & R \\
\end{bmatrix}$

\section{Nonlinear MPC}
\label{Sec:NMPC}
Trong chương này, chúng tôi trình bày chi tiết bộ điều khiển Nonlinear MPC thời gian liên tục cho quadrotor theo dõi quỹ đạo dựa trên vấn đề điều khiển tối ưu (OCP).


Hệ thống động học, động lực học được chúng tôi trình bày trong phần \ref{sec:problemstatement} từ đó đưa ra định nghĩa của vector trạng thái $\mathbi{x} \in \mathbb{R}^{12}$ như phương trình \ref{eq:stateEq} và vector đầu vào $\mathbi{u} \in \mathbb{R}^{4}$ theo phương trình \ref{eq:inputEq}:
\begin{equation}
    \mathbi{x} = \{\mathbi{\xi}^T, \mathbi{\eta}, \dot{\mathbi{\xi}}^T, \dot{\mathbi{\eta}}^T\}.
\label{eq:stateEq}
\end{equation}
\begin{equation}
    \mathbi{u} = \{\omega_1^2, \omega_2^2, \omega_3^2, \omega_4^2\}.
    \label{eq:inputEq}
\end{equation}

Tương tự như bộ điều khiển linear MPC, chúng tôi xây dựng bộ OCP phi tuyến như sau:
\begin{equation}
\min_{\mathbi{u}} \int_{t = 0}^{T} \bigg( \|^e\mathbi{x}(t) \|^2_{\mathbi{Q}} + \| \mathbi{u}(t)\|^2_{\mathbi{R}} \bigg) dt + \|  ^e\mathbi{x}(T)\|^2_{\mathbi{P}}
\label{eq:costLMPC}
\end{equation}

phụ thuộc vào:
\begin{equation}
\left\{\begin{matrix}
\dot{\mathbi{x}} = \mathbi{f}(\mathbi{x}, \mathbi{u})\\ 
\mathbi{u}_{min} \leqslant \mathbi{u}(t) \leqslant \mathbi{u}_{max} \\
\mathbi{x}_0 = \mathbi{x}(t_0), 
\end{matrix}\right.,
\end{equation}

trong đó, $\| \cdot \|$ biểu thị khoảng cách Euclidean. $\mathbi{Q} \geqslant 
 0$, $\mathbi{R} \geqslant 0$, $ \mathbi{P} \geqslant  0$ lần lượt là ma trận trọng số của trạng thái, đầu vào và trạng thái cuối cùng. T chiều dài đường dự đoán. $\| \mathbi{u}(t)\|^2_{\mathbi{R}}$ là giá trị đầu vào,  $\| ^e\mathbi{x}(T)\|^2_{\mathbi{P}}$  đánh giá độ lệch so với trạng thái mong muốn ở cuối đường dự đoán .$\|^e\mathbi{x}(t) \|^2_{\mathbi{Q}}$ là giá trị trạng thái được định nghĩa bằng sai số của trạng thái hiện tại $\mathbi{x}(t)$ và trạng thái tham chiếu $^r\mathbi{x}(t)$:
 \begin{equation}
     ^e \mathbi{x}(t) = \mathbi{x}(t) - ^r \mathbi{x}(t)
 \end{equation}

Bộ điều khiển thực hiện tối ưu hóa theo kiểu đường chân trời lùi dần. Do giới hạn về mặt tính toán, xử lý, yêu cầu giải quyết vấn đề (\ref{eq:costLMPC}) theo thời gian thực được đẩy lên hàng đầu. Nhiều kỹ thuật như \textit{Multiple shooting} được triển khai nhằm giải quyết yêu cầu trên. Bên cạnh đó, các ràng buộc về tính liên tục được áp đặt vào trong hệ thống động lực học và rời rạc hóa thời gian $t_0, t_1 \cdots, t_N$ trong khoảng $[t_k, t_{k+1}]$.

\section{KẾT QUẢ}
\label{Sec:KetQuaMoPhong}
\subsection{Thiết lập môi trường}
Mô hình robot được mô tả trong \ref{sec:problemstatement} với hàm chi phí trình bày trong phần \ref{Sec:NMPC} và \ref{Sec:LMPC}. Chúng tôi giới hạn tất cả các tham số đầu vào điều khiển theo biên độ sau:
\begin{equation}
    \mathbb{U} = \left\{ \mathbi{u} \in \mathbb{R}^4 | \left[ \begin{matrix}
0 \ (rad/s)\\ 
0 \ (rad/s)\\ 
0 \ (rad/s) \\
0 \ (rad/s)
\end{matrix} \right] \leqslant \mathbi{u } \leqslant \left[ \begin{matrix}
10 \ (rad/s)\\ 
10 \ (rad/s)\\ 
10 \ (rad/s) \\
10 \ (rad/s)
\end{matrix} \right] \right\}
\end{equation}

Bên cạnh đó, tỷ lệ thay đổi của tham số đầu vào điều khiển được giới hạn nhằm ngăn chặn các chuyển động đột ngột với:
\begin{equation}
\delta\mathbi{u }_{max} = - \delta\mathbi{u }_{min}  = \left[ \begin{matrix}
2 \ (rad/s)\\ 
2 \ (rad/s)\\ 
2 \ (rad/s) \\
2 \ (rad/s)
\end{matrix} \right] 
\end{equation}

Chúng tôi chọn ma trận trọng số như sau:
\begin{equation}
\begin{aligned}
    \mathbi{P} &= \mathbi{Q} = diag(\begin{bmatrix}
    ones(1,6) & zeros(1, 6)
    \end{bmatrix}) \\
    \mathbi{R} &= diag(\begin{bmatrix}
    0.1 & 0.1 & 0.1 & 0.1
    \end{bmatrix})
    \end{aligned}
\end{equation}

Bên cạnh đó, số bước dự đoán (prediction horizon) và số bước điều khiển (control horizon) được đặt tương ứng với $N = 18$, $N_u = 2$. 

Tiếp theo, vị trí khởi tạo cho quadrotor là $\mathbi{\xi} (0) = (0, 0, 0) ^T$ và vận tốc góc ban đầu là $\mathbi{\eta} (0) = (0, 0, 0) ^T$. Cuối cùng, quỹ đạo tham chiếu được thiết kế như sau:
\begin{equation}
    \begin{aligned}
        x &= 2 \ cos(\frac{2}{5}t) \ \ (m)  \\ 
        y &= 2 \ sin(\frac{2}{5}t)\ \  (m) \\
        z &= 0.2\ t \ \ \ \ \ \ \ \ (m)
    \end{aligned}
\end{equation}

\begin{table*}[ht]
\caption{Kết quả sai số bám quỹ đạo}
\centering
\begin{tabular}{m{0.06\textwidth} m{0.08\textwidth} m{0.08\textwidth} m{0.08\textwidth} m{0.1\textwidth} m{0.11\textwidth} m{0.1\textwidth} m{0.06\textwidth} m{0.07\textwidth} }
\hline
RMSE & Trục x (m)& Trục y (m) & Trục z (m) & Góc roll (rad) &  Góc pitch (rad) & Góc yaw (rad)  & $xyz$ (m) & $\phi \theta \psi$ (rad) \\ \hline
PD & 0.21120 & 0.04770 & \textbf{0.00032} & 0.00160 & 0.00270  & 0.00022 & \textbf{0.21650} & 0.00310 \\ \hline
BSC & \textbf{0.14330} & \textbf{0.01520} & 6.19800 & 0.19970 & 0.29290 & 0.00003  & 6.19970 & 0.35450 \\ \hline
SMC &  0.18820 & 0.02130 & 0.17990 & \textbf{0.00560} & 0.00240 & \textbf{0.00002}  & 0.26120 & 0.00620 \\ \hline
MPC &  0.23860 & 0.01840 & 0.00890 & 0.00062 & \textbf{0.00220} & 0.00019 & 0.23950 & \textbf{0.00230}\\\hline 
NMPC &  0.23880 & 0.01650 & 0.00200 & 0.00120 & \textbf{0.00220} & 0.00005 & 0.23940 & 0.00250 \\
\hline
\end{tabular}
\label{Tab:resultsystem}
\end{table*}

\subsection{Kết quả mô phỏng}
Trong phần này, các mô phỏng MATLAB được thực hiện để cho thấy hiệu quả của bộ điều khiển trong việc theo dõi quỹ đạo.
\begin{figure}[!ht]
    \centering
    \includegraphics[width=0.5\textwidth]{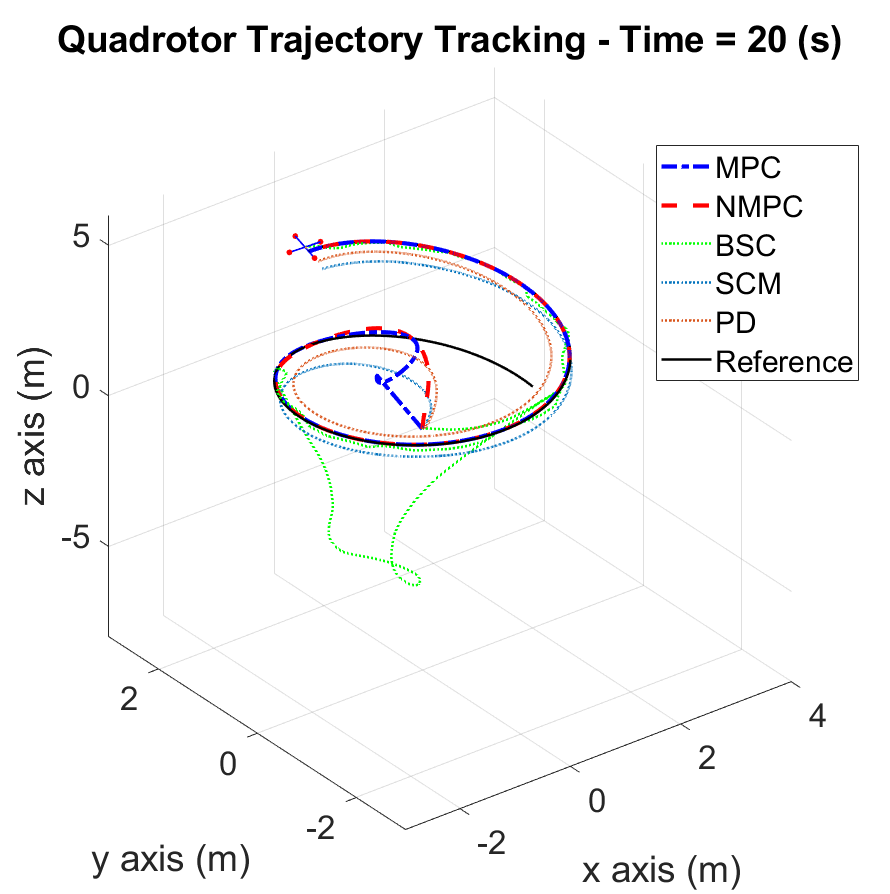}
    \caption{Các quỹ đạo bám của các bộ điều khiển}
    \label{fig:quadrotor_traj}
\end{figure}

Hình \ref{fig:quadrotor_traj} cho thấy các quỹ đạo được theo dõi của các bộ điều khiển trong đó bao gồm bộ điều khiển tuyến tính (Proportional Derivative - PD), phi tuyến (Sliding mode control - SCM, Backstepping control - BSC) và điều khiển tối ưu hóa (Linear MPC và Non Linear MPC). Tại thời điểm bắt đầu chuyển động, vị trí robot cách xa quỹ đạo do đó các bộ điều khiển có nhiệm vụ hướng về quỹ đạo tham chiếu. Dựa trên kết quả có thể nhận thấy rằng bộ điều khiển Non-Linear MPC cho khả năng theo dõi tốt  nhất.

\begin{figure*}[!ht]
    \centering
    \includegraphics[width=0.31\textwidth]{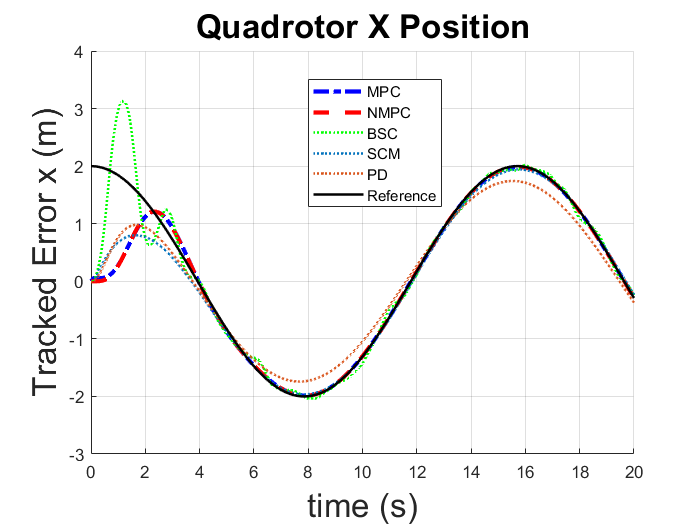}
    \includegraphics[width=0.31\textwidth]{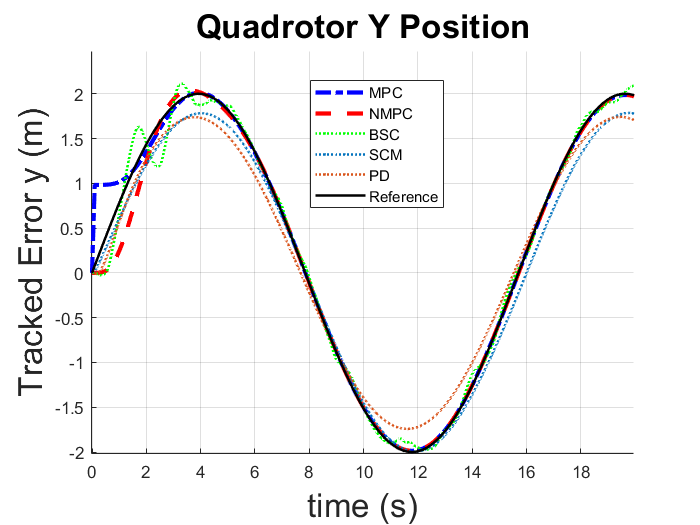}
    \includegraphics[width=0.31\textwidth]{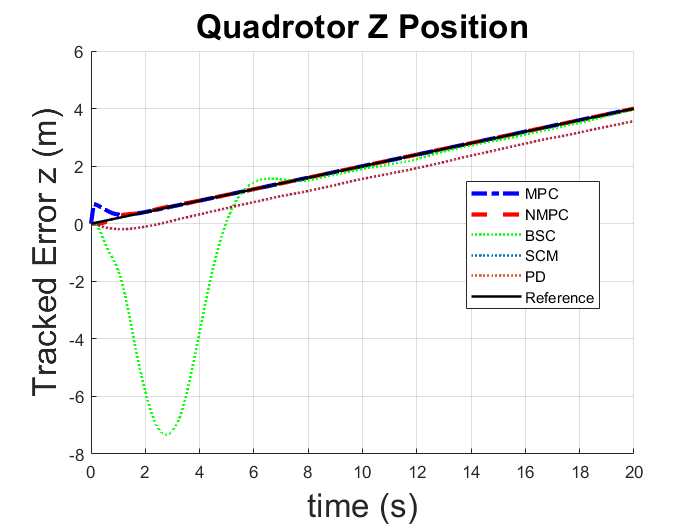}
    \includegraphics[width=0.31\textwidth]{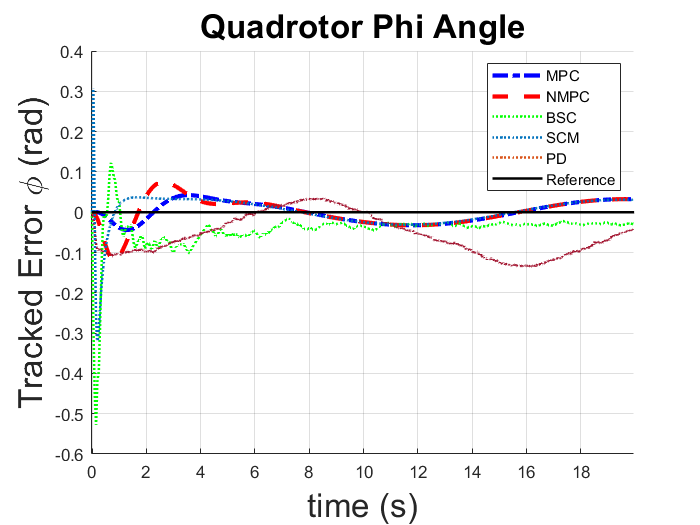}
    \includegraphics[width=0.31\textwidth]{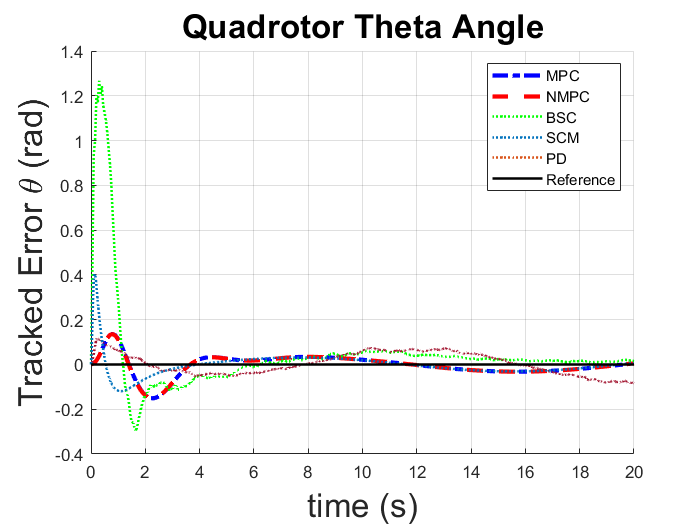}
    \includegraphics[width=0.31\textwidth]{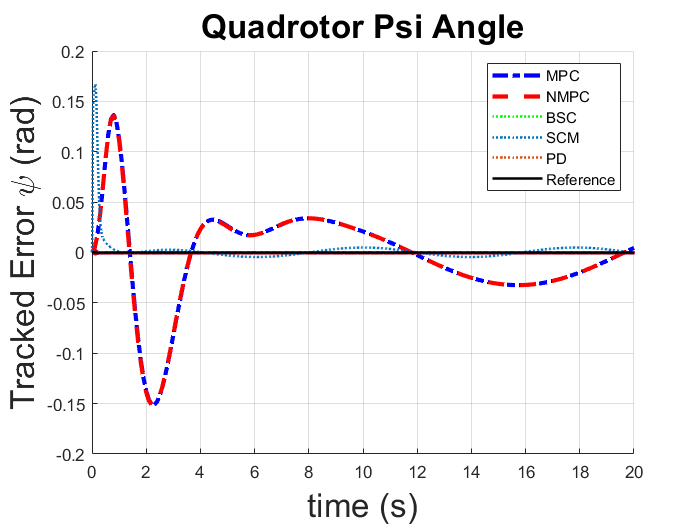}
    \caption{Sai số bám quỹ đạo theo từng thành phần}
    \label{fig:errorTrj}
\end{figure*}

Sai số theo dõi được hội tụ về 0, được thể hiện trong Hình \ref{fig:errorTrj}. Khi bắt đầu chuyển động, sai số là đáng kể do các trạng thái ban đầu và các ràng buộc về vận tốc, và theo thời gian sai số dần hội tụ về không. Bộ điều khiển None-Linear MPC có tốc độ hội tụ về 0 nhanh nhất (khoảng 5 giây với vị trí và 7 giây với góc) tiếp sau đó là Linear MPC từ đó cho thấy chất lượng theo dõi quỹ đạo tốt hơn ở bộ điều khiển Linear MPC và None-Linear MPC. Bên cạnh đó, bộ điều khiển None-Linear MPC không xuất hiện tín hiệu vọt lố (overshoot). Sai số cụ thể theo từng thành phần $(xyz)$ và $(roll, pitch, yaw)$ được thể hiện qua Bảng \ref{Tab:resultsystem}. Sai số khoảng cách và sai số góc được tính toán thông qua RMSE (Root-Mean-Square Error) của vị trí và góc. Kết quả cho thấy, sai số của bộ điều khiển MPC rất thấp (nhỏ hơn 0.24 (m) đối với sai số vị trí và 0.0026 (rad) đối với sai số góc) từ đó đảm bảo độ chính xác, ổn định trong quá trình bám quỹ đạo. Tuy nhiên trong mô phỏng, bộ điều khiển PD vẫn cho thấy hiệu quả tốt nhất đối với sai số vị trí.

\begin{figure*}[!ht]
    \centering
    \includegraphics[width=0.23\textwidth]{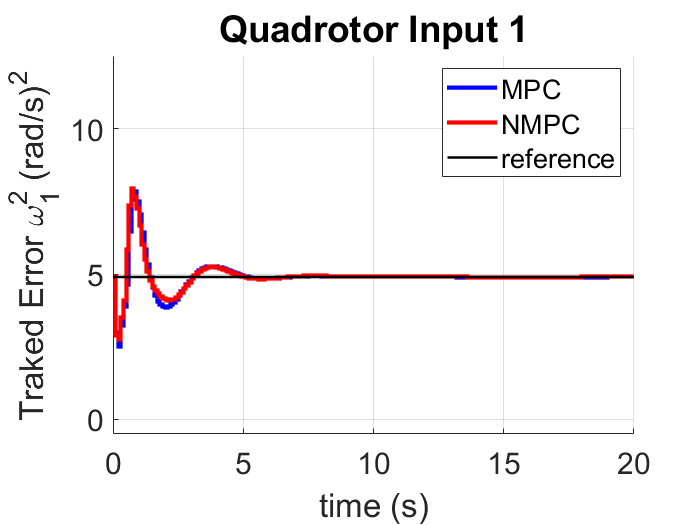}
    \includegraphics[width=0.23\textwidth]{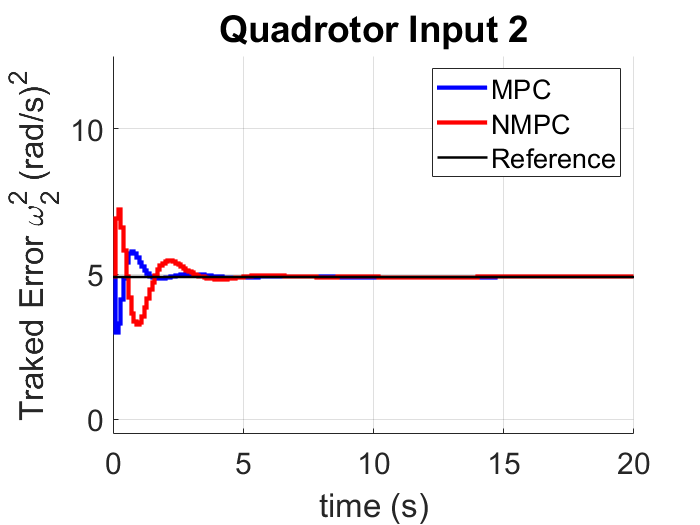}
    \includegraphics[width=0.23\textwidth]{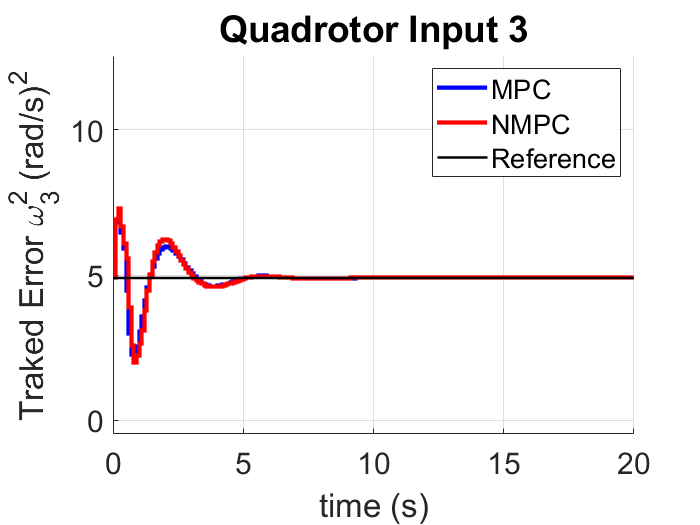}
    \includegraphics[width=0.23\textwidth]{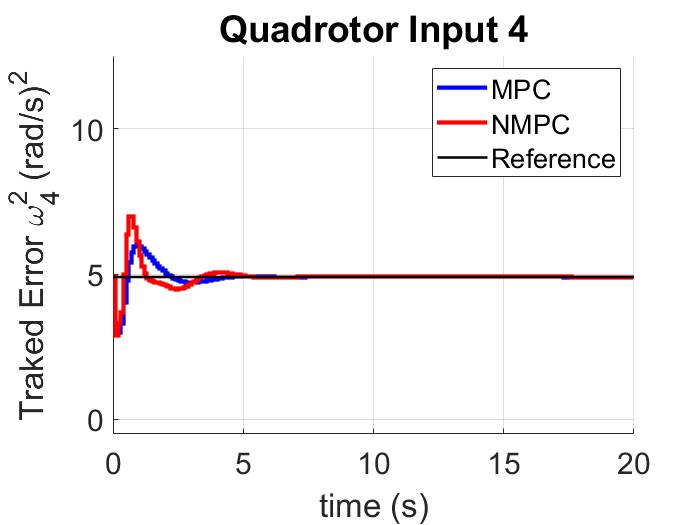}
    \caption{Sai số bộ tín hiệu điều khiển}
    \label{fig:inputState}
\end{figure*}
Trạng thái tín hiệu điều khiển được thể hiện trong Hình \ref{fig:inputState} Các đầu vào điều khiển được điều khiển đến giá trị mục tiêu là 4,9 $(rad/s)^2$ trong khoảng 10 giây.

\section{KẾT LUẬN}
\label{Sec:KetLuan}
Trong bài báo này, chúng tôi đã trình bày bộ điều khiển dự đoán mô hình tuyến tính và phi tuyến để theo dõi quỹ đạo của Quadrotor. Mô hình động học, động lực học của quadrotor được cũng cấp từ đó chứng minh tính hiệu quả trong việc theo dõi của cả hai bộ điều khiển. Kết quả mô phỏng triển khai trên phần mềm mô phỏng Matlab nhằm so sánh độ chính xác, tính hội tụ, ổn định của Linear MPC và Non-Linear MPC so với các bộ điều khiển tuyến tính, phi tuyến khác. 
%

\bibliographystyle{IEEEtran}
\balance
\bibliography{reference}
\end{document}